# Cost and Actual Causation

Liang Zhou
zhouliang@pku.edu.cn

**Abstract**

I propose the purpose our concept of actual causation serves is minimizing various cost in intervention practice. Actual causation has three features: non-redundant sufficiency, continuity and abnormality; these features correspond to the minimization of exploitative cost, exploratory cost and risk cost in intervention practice. Incorporating these three features, a definition of actual causation is given. I test the definition in 66 causal cases from actual causation literature and show that this definition' application fit intuition better than some other causal modelling based definitions.

## 1 Introduction

In the literature of actual causation, a question has been frequently asked: why do we possess this particular concept of actual causation? what kind of purpose does the concept of actual causation serve? Hitchcock and Knobe proposed that "people's concept of actual causation enables them to design effective interventions "[1], i.e., actual causation concept help people to select the preferable intervention which

[1] Hitchcock, C. and Knobe, J. [2009]: 'Cause and norm', Journal of Philosophy, 106,p.590.

"works by targeting abnormal aspects of the situation and replacing them with more normal ones".[2]

In this paper I offer a somehow different view of what kind of interventions are preferable: preferable interventions should satisfy non-redundant sufficiency, continuity and abnormality, because in satisfying these features the cost in intervention practice is qualitatively minimized. In the following part of introduction section I will explain more about cost in intervention practice.

In order to achieve some desired goal in real life, we need to make intervention or prediction of whether some previous events occurs or not. There are two types of these intervention/prediction activities: exploitative ones and exploratory ones.

No matter whether the goal-pursuing activity is exploitative or exploratory, generally the activity would have practical costs, i.e., certain amount of limited resource would be consumed during the activity. There are at least four kind of such practical cost.[3]

(1) Exploitative Cost. Sometimes the goal-pursuing activity is exploitative. People manipulate the presence or absence of some previous events to ensure the presence or absence of a following event. People collect information of the presence or absence of some previous events to predict the presence or absence of a following event. For instance, suppose I am to cook an egg for breakfast. I put an egg into boiling water and wait to obtain a boiled egg. I observe the flame color and timing to predict how tender the cooked egg will be. In short, in order to ensure the occurrence of an exploitative goal event, the agent should intervene with some cause-like event of the goal event, and the intervention activities have practical cost. For instance, in order to obtain a boiled egg, I do intervention with the egg, and the intervention activity consumes some electric energy, water and time.

[2] Ibd., p.612

[3] In the following sections I will only discuss intervention activities and their relationship to causality. I will not discuss prediction activities.

(2) Exploratory Costs. Sometimes the goal-pursuing activity is exploitative. The goal of the activity is not to achieve a concrete physical outcome such as a cooked egg. Rather, the goal is to explore the world and extend our causal knowledge, to know more about which event causes some other event. For instance, I may do a serial of experiment to find the shortest time required to boil an egg. In this case, the goal of my experiments is not to cook a particular egg, but to explore the causal relationship between the egg boiling process and the processed egg. In short, in order to achieve some exploratory goals, the agent should run some actual or virtual experiment trials, and these experimental activities have practical cost. For instance, the ‘fastest egg-boiling’ experiment above cost electric energy, water and time. Even if the agent merely computes the shortest egg boiling time theoretically but doesn’t actually run the experiment, the theoretical process would cost some computing resource and time.

(3)Risk Cost: When we use a specific set of intervention to promote some goal event, there is risk that the goal event failed to occur even if the intervention is done. For instance, suppose I boil an egg for two minutes and pick it out, it is possible that the egg is still in semi-liquid state and uneatable. Then, the bad possibility that the boiled egg is uneatable is the risk cost I have to pay for the intervention of boiling the egg for two minutes.

(4) Normative Costs. If the intervention activity violate some practical/moral norm, the norm-violation aspect of the activity can be seen as the normative cost of the activity. Suppose My neighbor Jack and I share a kitchen. Only two eggs are left in the kitchen and they all belong to Jack. In the morning Jack boiled one for breakfast and I boiled the other without notifying jack. If we ask what causes the fact that Jack has no egg to use for lunch, we are more inclined to say my action in the morning rather than Jack's is the cause. The reason for this is that my egg-using activity violate a practical norm ("Not to use others' property without permission") and my neighbor's activity doesn't. In this paper I will not further address the issue of normative cost ,but simply assume that normative cost can be treated like risk cost is treated.

Because of the resource-limited nature of the real world, we have good reason to minimize the four kind of cost above. The cost minimization can be done either qualitatively or quantitatively.

On the qualitative level, the question is whether a particular set of event is both efficient and nonredundant in promoting a following event. Being efficient means intervention with this set of events somehow ensure the following event's occurrence. Being nonredundant means that intervention with any subset of this event set doesn't ensure the following event as good as the original set.

On the quantitative level, for any two efficient and nonredundant intervention set $\vec{A}$ and $\vec{B}$ for some goal event e, if quantitative information of all relevant cost of both sets is available, the agent can quantitively compare $\vec{A}$ and $\vec{B}$'s total cost and choose the one with less cost.

In most real scenarios we don't have detailed quantitative knowledge of the relevant intervention cost above. So epidemically what is more relevant to the concept of actual causation is the question of cost minimization on the qualitative level.

I propose that the question of  qualitatively minimize intervention cost is closely linked to the concept of actual causality. I suggest that for a goal event Y = y, if an event set $\vec{X} = \vec{x}$ ensures Y=y with qualitative minimal cost in the four aspect above, then $\vec{X} = \vec{x}$ is a actual cause set for Y=y, and any conjunct X=x of $\vec{X} = \vec{x}$ is an actual cause of Y=y. Speaking of counterfactual scenarios, event set $\vec{X} = \vec{x}$ is actual cause set of event Y = y in scenario S if and only if, in all scenarios similar to S in structure, $\vec{X} = \vec{x}$ ensures event E = e and the cost of the intervention with $\vec{X} = \vec{x}$ is qualitatively minimal in the four aspect mentioned above.

In the following chapters, I will further elaborate this idea. In Chapter 2, I introduce formalization in causality modeling and define what is a scenarios similar to the actual scenario S.

In Chapter 3, I discuss the Mackie[4] intuition that if event X = x causes Y = y, then X=x should be a member of a nonredundant sufficient condition set of Y. I will show that this nonredundant sufficiency requirement is related to the minimization of exploitative cost.

In Chapter 4, I discuss the intuition that if event X = x causes Y = y, then there should be some specific causal chain between X=x and Y=y. I will show that this chain requirement is related to the minimization of exploratory cost.

In Chapter 5, I discuss the intuitive distinction between an event’s key cause and its background conditions, i.e., the intuition behind causal selection which render causation somehow inegalitarian. I will show that this intuition is related to the minimization of risk cost.

---

[4] Mackie, J.L.[1965]:'Causes and conditions’, American Philosophy Quarterly 2/4, 261-264. Reprinted in E.Sosa and M. Tooley(Eds.), Causation, Oxford University Press, 1993.

In Chapter 6, I propose an definition of actual causation which combines the analysis of its previous chapters. I will apply this definition, Halpern and Hitchcock (2015) definition and Weslake (2015) definition to 66 causal cases, and show that my definition fit intuition in more cases than others.

# 2 Causal Modelling

Following Pearl and Halpern's 2005 paper[5], I adopt causal modeling approach and define some concepts formally in this chapter.

I limit the discussion in deterministic scenarios and don't consider genuine randomness/ quantum randomness. I first define a Universal Causal Model that models the entire world where the actual causal scenario is embedded in.

**Definition 2.1**, Universal Causal Model. A universal causal model $M_w$ is a pair $((\overrightarrow{V_w}, R_w), F_w)$. Here, $\overrightarrow{V_w}$ is the set of all event variables mutually logically independent in the actual world. $R_w$ is a function that determines the range of values for the variables in $\overrightarrow{V_w}$, which associates each variable X in $\overrightarrow{V_w}$ with a range of values $R_w$ (X). $F_w$ is a function that assigns structural equations to variables, which maps each variable X in $\overrightarrow{V_w}$ to an function $f_{w_x}(\overrightarrow{PA_x})$ ; $\overrightarrow{PA_x}$ is the markovian parents set [6] of X and $\overrightarrow{PA_x} \subseteq \overrightarrow{Vw}$ -{X}; function $f_{w_x}$ maps each value setting $\overrightarrow{pa_x}$ of $\overrightarrow{PA_x}$ to a value x ∈Rw(X).

[5] Halpern, J. Y. and Pearl, J. [2005]: 'Causes and explanations: A structural-model approach. Part I: Causes', British Journal for Philosophy of Science, 56, pp. 843–87.

[6] According to Pearl, J. [2000]: Causality: Models, Reasoning, and Inference, NY: Cambridge

The Universal Causal Model contains the structural relationships between all the event variables in a world. We can extract a simplified model M from $M_w$ , which only preserves the events relevant to a particular effect event E=e. Hence we can define a Simplified Model concerning E=e.

**Definition 2.2**, Simplified Model. A simplified model $M = \left((\vec{V}, \vec{U}, \vec{T}, R), F\right)$ for event E = e is a pair in which, (1) $\vec{V} \subseteq \overrightarrow{V_w}$ is the set of endogenous variables, (2) $\vec{U} \subseteq \overrightarrow{V_w} - \vec{V}$ is the set of exogenous variables whose value is independent from $\vec{V}$ 's value. (3) $\vec{T}$ is the set of dependant variables whose value depent on value of some variable in $\vec{V}$ but are not considered endogenous by the modeler.(4) R is a function that determines the range of values for the variables in $\vec{V}$ , which maps each variable X in $\vec{V}$ to a range of values $R_w$ (X).(0) F is a function that assigns structural equations to variables, which maps each variable X in $\vec{V}$ to an function $f_{w_x}(\overrightarrow{PA_x})$ ; (6) function $f_{w_x}$ maps each value setting $\overrightarrow{pa_x}$ of $\overrightarrow{PA_x}$ to a value x ∈R(X), as it does in the universal causal model.

A scenario concerning event E=e is a pair ($\vec{u}$,M), where M is a simplified model for E=e, and $\vec{u}$ is a value setting for the exogenous variables $\vec{U}$ in M.

I completely follow Halpern&Hitchcock's work in signifying intervention and formula's truth value in a model:

Intervention with variables in a causal model: "Setting the value of some variable X to x in a causal model M=(S,F) by means of an intervention results in a new causal

---

University Press. P.14, "let V = {X], . . . , Xn } be an ordered set of variables, and let P(v) be the joint probability distribution on these variables. A set of variables PA j is said to be Markovian parents of Xj if PAj is a minimal set of predecessors of Xj that renders Xj independent of all its other predecessors."

model denoted $M_{\vec{X}=\vec{x}}$ . $M_{\vec{X}=\vec{x}}$ is identical to M, except that the equation for X in F is replaced by X=x."[7]

Causal formula's truth value in a causal model: "A causal formula φ is true or false in a causal model, given a context. We write $(\vec{u}, M) \models \varphi$ if the causal formula ' is true in causal model M given context $\vec{u}$. The $\models$ relation is defined inductively.

$(\vec{u}, M) \models X = x$ if the variable X has value x in the unique solution to the equations in M in context $\vec{u}$ (that is, the unique vector of values for the endogenous variables that simultaneously satisfies all equations in M with the variables in $\vec{U}$ set to $\vec{u}$). The truth of conjunctions and negations is defined in the standard way. Finally, $(\vec{u}, M) \models [Y = y]\varphi$ if $(\vec{u}, M_{\vec{X}=\vec{x}}) \models \varphi$. "[8]

Furthermore, we can partition the endogenous variable set $\vec{V}$ into the initial variable set $\overrightarrow{V_{ini}}$ and the non-initial variable set $\overrightarrow{V_{inter}}$ : For any variable X, X is an initial variable iff the arguments set of the structural function $f_{wx}$ of X has no intersection with $\vec{V} \cup \vec{T}$. In other words, X is an initial variable iff X's value doesn't depend directly or indirectly on other variable's value in $\vec{V}$.

Only when the value setting of $\vec{U}$ is within a certain range can it ensure that the condition (3) of the simplified model definition is satisfied and that the simplified model M is locally deterministic. [9] We can call this specific range of value settings of $\vec{U}$ its deterministic range.

[7] Halpern, J. Y. and Christopher Hitchcock. [2015]: 'Graded Causation and Defaults', British Journal for Philosophy of Science, 66, p.419.

[8] Ibid., p.422.

**Definition 2.3**, Deterministic Range of $\vec{U}$ . For any subset $\vec{A}$ of $\overrightarrow{V_w}$ , its value setting range is Range($\vec{A}$) = $\times_{V\in\vec{A}}$ $R_w(V)$. A subset $\mathrm{Range_d}(\vec{U})$ of Range($\vec{U}$) is a deterministic range concerning ($\vec{u}$,M), iff for any x ∈ Range($\overrightarrow{V_{inter}\cup T}$), any $\overrightarrow{u_1}$ ∈ Range'($\vec{U}$) , any $\vec{v}$ ∈ Range($\vec{V}$ ), and any $\vec{t}$∈ Range($\vec{T}$ ), $f_{w_x}(\overrightarrow{u_1},\vec{v},\vec{t}) =$

$f_{wx}(\vec{u},\vec{v},\vec{t})$.

In other words, whenever $\overrightarrow{u_1}$∈ Range'($\vec{U}$), the values of all non-initial endogenous variables and dependent variables are only influenced by other endogenous variables and dependent variables, but not influenced by $\vec{U}$'s value change inside Range'($\vec{U}$). Because this deterministic feature, We can fix these non-initial variable and dependent variable's structural function's exogenous arguments into some constant value, so that these structure functions don't contain exogenous variables any more.

With the concept of Deterministic Range, we can define two different criteria for "what is a causal scenario similar to scenario S ":

**Definition 2.4**, Reliable Similar Scenario Set. For scenario S= ($\vec{u}$,M), set {($\overrightarrow{u'}$,M) |$\overrightarrow{u'}$ ∈ $\mathrm{Range_d}(\vec{U})$ } is a reliable similar scenario set to S.

**Definition 2.5**. General Similar Scenario Set. For scenario S= ($\vec{u}$,M), scenario S'=($\overrightarrow{u'}$, M') is a general similar scenario to S, iff $\vec{U} = \vec{U}', \vec{V} = \vec{V}', \vec{T} = \vec{T}'$. All such scenarios constitute a general similar scenario set of S.

---

[9] See Halpern, J. Y. and Pearl, J. [2005]: 'Causes and explanations: A structural-model approach. Part I: Causes', British Journal for Philosophy of Science, 56, p.848.

The two criterions for scenario similarity correspond to our two understanding of scenario similarity:

In Reliable Similar Scenarios of the actual scenario, the value setting of exogenous variables is sufficient to ensure that each endogenous variable acts according to its structural equation. We can regard the whole causal structure as a reliable machine which can process input and produce output regularly. For instance, suppose an agent is to use a new reliable typewriter to print the word "hello" on a piece of paper. The agent only needs to hit characters “hello” on the keyboard without worrying about the internal process of the typewriter, because he/she has good reason to assume the typewriter’s internal process will work precisely according to its structural functions which are designed for typing-printing.

In General Similar Scenarios of the actual scenario, the value setting of exogenous variables is insufficient to ensure that each endogenous variable acts according to its structural equation. If the agent only know that a given scenario is generally similar to the actual scenario, but has no further knowledge to assure that the scenario is reliable similar to the actual scenario, then the agent needs to intervene with some endogenous variables to ensure that the variable follows their structural equation. We can regard the whole causal structure in the scenario as an worn-out machine whose internal parts do not necessarily follow the prescribed rules. In order to obtain desirable outcome with this machine, the agent needs not only to give correct input on the input-interface, but also to intervene with the internal parts to ensure they work appropriately. For instance, suppose the actor is to use a worn-out typewriter to print the word "hello" . He/she has to both hit “hello” on the keyboard and take some maintenance operations to ensure the typewriter work well.

# 3 Exploitative Cost and Sufficiency Requirement of Causation

With the causal modelling support of Chapter 2, we can investigate actual causation and start with the simplest kind of causal scenario.

**Definition 3.1**, Direct Connection Model. A simplified model M is a direct connection model of event E=e, iff M's endogenous variable set consists only of some event E=e and its structural function's argument set $\overrightarrow{V_E} = \{X_1, \ldots, X_n\}$ such that: (1) E= fE($\vec{U}$, $X_1$, $\ldots X_n$); (2) for any $X_i$ in $\overrightarrow{V_E}$, $X_i = \text{fxi}(\vec{U})$. In other words, in this model, all endogenous variables other than E are directly linked to E through the structural function of E, and they are only affected by exogenous variables.

For instance, Jack's flowers in the backyard survive iff Jack waters the flowers and there is no hail. I.e., Flower = Water & ~ Hail. Here, the three variables Water, Hail and Flower and the relationship between them constitute a direct connection model.

Suppose ($\vec{u}$, M) is an actual direct connection scenario and ($\vec{u}$, M) $\models$ E=e. When the agent faces a similar situation ($\overrightarrow{u'}$, M), if the actor is to ensure the event E = e and to minimize cost , then it would be reasonable to take the following intervention strategy: Only intervene with a subset $\vec{S} = \vec{s}$ of E's structural function's arguments set, and $\vec{S}$ should satisfies two conditons: (1) intervention with $\vec{S} = \vec{s}$ ensures E=e, no matter how other variable's value changes.(2) Any subset of S does not satisfy condition (1).

The condition (1) above ensures that the strategy is efficient in producing E=e, and the condition(2) ensures that the strategy is nonredundant. Based on this strategy, we can define direct cause of a event.

**Definition 3.2** Direct Cause Preliminary. In scenario ($\vec{u}$, M), event C = c is direct cause of event E = e, iff C = c belongs to a set $\vec{S} = \vec{s}$ and $\vec{S} = \vec{s}$ satisfies: (1) ($\vec{u}$, M) ⊨ ($\vec{S} = \vec{s}$ & E=e); (2)let $\vec{W} = \vec{V}$-$\vec{S}$-{E},then for any $\vec{W} = \vec{w}'$, ($\vec{u}$ ,M) ⊨ [$\vec{W} = \vec{w}'$] E=e; (3) no subset of $\vec{S} = \vec{s}$ satisfies both (1)and (2). We can call the set $\vec{S} = \vec{s}$ which satisfies (1)(2)(3) a direct cause set of E=e.

In scenarios more complex than direct connection scenarios, some events are indirectly connected to the effect event. Then, the question of how to determine a reasonable intervention strategy will depend on which kind of similar scenarios are relevant to the actual scenario and worth consideration. Following this distinction of the two criteria for scenario similarity, we can distinguish two kind of intervention strategy:

(1) Intervention in unreliable similar scenarios to the actual scenario. In this type of scenarios, the value setting of the exogenous variable is insufficient to ensure that each endogenous variable follows its structured equation. The agent needs to intervene with some endogenous variables to ensure these variables follow their structural equation.

This leads to the idea that there are two kind of intervention with an individual variable: value intervention and function intervention. In value intervention, the agent fixes the value of event variable X to X = x by intervention. In function intervention, the actor ensures that X's value assignment follows its structural function $f_x$. For instance, in the worn-out typewriter case above, the agent should do both value intervention (e.g., typing on keyboard) and function intervention (e.g., do some internal maintenance to make the typewriter work well) to ensure his/her desired outcome.

We can use $\overrightarrow{S_v}$ to denote the set of events that the agent does value intervention with, use $\overrightarrow{S_f}$ to denote the set of events the agent does function intervention with, and use $Loss_v$ (X = x) and $Loss_f$ (X = x) to denote the cost of value intervention and function intervention for event X = x. Then we can formulate the equation for an exploitative intervention strategy's total intervention costs:

$$\mathrm{Loss}_{\mathrm{intervention}} = \sum_{Xi\ in\ \overrightarrow{S_v}} Loss_v(X_i = x_i) \ + \ \sum_{Xj\ in\ \overrightarrow{S_f}} Loss_f(X_j = x_j)$$

If the agent wants to both ensure E=e and to minimize the intervention cost, adopting the intervention strategy below will be reasonable:

**Intervention Strategy in Unreliable Scenario**: Do value intervention to some $\overrightarrow{S_v} = \overrightarrow{s_v}$ and do function intervention to some $\overrightarrow{S_f}$ , while $\overrightarrow{S_v} = \overrightarrow{s_v}$ and $\overrightarrow{S_f}$ should satisfy: (1) $\overrightarrow{S_v}$ and $\overrightarrow{S_f}$ are subsets of the endogenous variables and mutually disjoint; (2) value intervention with $\overrightarrow{S_v} = \overrightarrow{s_v}$ and function intervention with $\overrightarrow{S_f}$ ensures E=e, no matter how other endogenous variables' value changes. (3) for any subset $\overrightarrow{S_v'} = \overrightarrow{s_v'}$ of $\overrightarrow{S_v}$ and subset $\overrightarrow{S_f'}$ of $\overrightarrow{S_f}$ , the pair ($\overrightarrow{S_v'} = \overrightarrow{s_v'}$ , $\overrightarrow{S_f'}$) doesnot satisfy (1) and (2).

The condition (1) of the strategy avoids inconsistent interventions; the condition (2) and (3) ensures that the strategy is both efficient and nonredundant in producing the effect event E=e. Based on this strategy, I propose the Sufficiency Requirement in Unreliable Scenario for actual causation:

**Proposition 3.3**, Sufficiency Requirement in Unreliable Scenario for actual causation. In an case where unreliable scenarios similar to ($\vec{u}$,M) are relevant, if event C = c is actual cause of E = e, then the scenario should satisfies: (1) exist $\overrightarrow{S_v} = \overrightarrow{s_v}$ and $\overrightarrow{S_f}$ such that for endogenous variable set's actual value setting $\vec{V} = \vec{v}$ , $\overrightarrow{S_v} = \overrightarrow{s_v} \subseteq \vec{V} =$

$\vec{v}$, $\overrightarrow{S_f} \subseteq \vec{V}$, and $\overrightarrow{S_v} \cap \overrightarrow{S_f}$ = {}, X=x $\in \overrightarrow{S_v} = \overrightarrow{s_v}$; (2) let $\vec{W} = \vec{V} - \overrightarrow{S_v} - \overrightarrow{S_f}$ -{E}, for any $\vec{W} = \vec{w}'$, ($\vec{u}$ ,M) ⊨ [$\overrightarrow{S_v} = \overrightarrow{s_v}$ , $\vec{W} = \vec{w}'$] E=e ;(3) for any subset $\overrightarrow{S_v'}$ of $\overrightarrow{S_v}$ and subset $\overrightarrow{S_f'}$ of $\overrightarrow{S_f}$ , the pair ($\overrightarrow{S_v'}$ , $\overrightarrow{S_f'}$) doesnot satisfy (1) and (2).

We can call a set $\overrightarrow{S_v} = \overrightarrow{s_v}$ which satisfies (1) and (2) in proposition 3.3 an sufficient set for E = e in unreliable scenarios similar to ($\vec{u}$, M).

(2) Intervention in reliable scenarios. In this type of scenarios, the value setting of the exogenous variable is sufficient to ensure that each endogenous variable follows its structural function.

In cases where this type of scenarios is relevant, a reasonable intervention strategy for the agent is: Do value intervention to some $\overrightarrow{S_v} = \overrightarrow{s_v}$ and do function intervention to some $\overrightarrow{S_f}$ , while $\overrightarrow{S_v} = \overrightarrow{s_v}$ and $\overrightarrow{S_f}$ should satisfy: (1) $\overrightarrow{S_v}$ and $\overrightarrow{S_f}$ are subsets of the endogenous variables and mutually disjoint; (2) value intervention with $\overrightarrow{S_v} = \overrightarrow{s_v}$ , function intervention with $\overrightarrow{S_f}$ and the following-structural-function nature of noninitial variables ensures E=e, no matter how other endogenous variables' value changes. (3) for any subset $\overrightarrow{S_v'} = \overrightarrow{s_v'}$ of $\overrightarrow{S_v}$ and subset $\overrightarrow{S_f'}$ of $\overrightarrow{S_f}$ , the pair ($\overrightarrow{S_v'} = \overrightarrow{s_v'}$ , $\overrightarrow{S_f'}$) doesnot satisfy (1) and (2).

Based on this strategy, I propose the Sufficiency Requirement In Reliable Scenario for actual cause:

**Proposition 3.4**, Sufficiency Requirement In Reliable Scenario for actual cause. In an case where unreliable scenarios similar to ($\vec{u}$,M) are relevant, if event C = c is actual cause of E = e, then the scenario should satisfies: (1) exist $\overrightarrow{S_v} = \overrightarrow{s_v}$ and $\overrightarrow{S_f}$ such that for endogenous variable set's actual value setting $\vec{V} = \vec{v}$ , $\overrightarrow{S_v} = \overrightarrow{s_v} \subseteq \vec{V} = \vec{v}$, $\overrightarrow{S_f} \subseteq \vec{V}$, and $\overrightarrow{S_v} \cap \overrightarrow{S_f}$ = {}, X=x $\in \overrightarrow{S_v} = \overrightarrow{s_v}$; (2) let $\vec{W} = \overrightarrow{V_{ini}} - \overrightarrow{S_v} - \overrightarrow{S_f}$ , for any $\vec{W} = \vec{w}'$, $(\vec{u}, M) \models [\overrightarrow{S_v} = \overrightarrow{s_v}, \vec{W} = \vec{w}']$ E=e ;(3) for any subset $\overrightarrow{S_v'}$ of $\overrightarrow{S_v}$ and subset $\overrightarrow{S_f'}$ of $\overrightarrow{S_f}$, the pair $(\overrightarrow{S_v'}, \overrightarrow{S_f'})$ doesnot satisfy (1) and (2).

We can call a set $\vec{S} = \vec{s}$ which satisfies (1) and (2) in proposition 3.4 an sufficient set for E = e in reliable scenarios similar to $(\vec{u}, M)$.

# 4 Exploratory Cost and Continuity Requirement of Causation

In this chapter I discuss the chain feature of actual causation and show that this chain feature helps minimize exploratory cost.

I mentioned in the introduction that sometimes the goal we seek is exploratory, i.e., the goal is to explore the world and extend causal knowledge. A straight way to carry out this exploration is to actually or virtually test different sets of events to see whether intervention with these event set can ensure the effect event no matter how other events' value settings change. However, these actual or virtual tests have cost, such as the exploitative cost of intervention in each trial and the risk cost that some undesired events occur.

The following three reasoning operation (extrapolation operation, interpolation operation and flank operation) remarkably reduce the number and cost of tests that the agent need to try for exploration. To understand the three principles, I first define the concept of cause net:

**Definition 4.1** Cause Net. For a scenario $(\vec{u}, M)$ concerning event E=e:

(1) any direct cause set $\vec{N} = \vec{n}$ of E=e is a cause net of E=e.

(2) for any cause net $\overrightarrow{N'} = \overrightarrow{n'}$ of E=e, replace some member X=x of $\overrightarrow{N'} = \overrightarrow{n'}$ with one of X=x's direct cause set, the new set is also a cause net of E=e.

It can be proved that if $\vec{\mathrm{N}} = \vec{\mathrm{n}}$ is a cause net of E=e in $(\vec{\mathrm{u}}, \mathrm{M})$, then $\vec{\mathrm{N}} = \vec{\mathrm{n}}$ is a sufficient set for E=e in reliable scenarios of $(\vec{\mathrm{u}}, \mathrm{M})$. In other words, if all noninitial event variables follow their structural functions, intervention with $\vec{\mathrm{N}} = \vec{\mathrm{n}}$ can ensure E=e no matter how other initial variables change. With the concept of cause net , I now introduce the three reasoning operations:

**Interpolation Operation**. Suppose $\vec{\mathrm{S}} = \vec{\mathrm{s}}$ is a cause net of E = e in scenario $(\vec{\mathrm{u}}, \mathrm{M})$ . Replace some member X=x of $\vec{\mathrm{S}} = \vec{\mathrm{s}}$ with a set $\vec{\mathrm{H}} = \vec{\mathrm{h}}$, and $\vec{\mathrm{H}} = \vec{\mathrm{h}}$ should satisfy: for each direct cause chain {X=x, …, E=e} where each chain member is a direct cause of its successor, the successor of X=x is included in $\vec{\mathrm{H}} = \vec{\mathrm{h}}$. This operation is a interpolation operation on $(\vec{\mathrm{S}} = \vec{\mathrm{s}}, X = x)$. It can be proved that the new set $\vec{\mathrm{S}'} = \vec{\mathrm{s}'}$ after replacement is an sufficient set in reliable scenarios of $(\vec{\mathrm{u}}, \mathrm{M})$.

**Extrapolation Operation**: Suppose $\vec{\mathrm{S}} = \vec{\mathrm{s}}$ is a cause net of E = e in scenario $(\vec{\mathrm{u}}, \mathrm{M})$ . Replace some member X=x of $\vec{\mathrm{S}} = \vec{\mathrm{s}}$ with a one of X=x's sufficient set $\vec{\mathrm{I}} = \vec{\mathrm{i}}$. This operation is a Extrapolation Operation on $(\vec{\mathrm{S}} = \vec{\mathrm{s}}, X = x)$. It can be proved that the new set $\vec{\mathrm{S}''} = \vec{\mathrm{s}''}$ after replacement is an sufficient set in reliable scenarios of $(\vec{\mathrm{u}}, \mathrm{M})$.

**Flank Operation**: Suppose $\vec{\mathrm{S}} = \vec{\mathrm{s}}$ is a cause net of E = e in scenario $(\vec{\mathrm{u}}, \mathrm{M})$ . Do a interpolation operation on $(\vec{\mathrm{S}} = \vec{\mathrm{s}}, X = x)$ and get a new set $\vec{\mathrm{S}'} = \vec{\mathrm{s}'}$, then do a extrapolation operation on $(\vec{S'} = \vec{\mathrm{s}'}, Z = z)$ for each Z$\in\vec{S'} - \vec{\mathrm{S}}$ . These serial of operations constitute a Flank Operation. the new set $\vec{\mathrm{S}'''} = \vec{\mathrm{s}'''}$ after operation is an sufficient intervention set in reliable scenarios of $(\vec{\mathrm{u}}, \mathrm{M})$.

We can call the average length of all direct causal chains from $\vec{\mathrm{S}} = \vec{\mathrm{s}}$ members to E = e as the distance from $\vec{\mathrm{S}} = \vec{\mathrm{s}}$ to E = e. If $\vec{\mathrm{S}} = \vec{\mathrm{s}}$ is a cause net of E=e, then a sufficient set less distanced from E=e than $\vec{S} = \vec{s}$ can be acquired by interpolation operation, a sufficient set more distanced from E=e than $\vec{S} = \vec{s}$ can be acquired by extrapolation operation, and a sufficient set as distanced from E=e as $\vec{S} = \vec{s}$ may be acquired by a flank operation. Hence by using the three operations above, an agent can extend his/her causal knowledge without actual or virtual experiment trials.

There is some special connection between the interpolation operation and chain of direct causes. By definition, only events which belong to a direct cause chain toward E=e can replace others or be replaced in the interpolation operation. This means only events in direct cause chain to E=e can be start point or end point in an interpolation operation. [10]

Therefore, the within-direct-chain feature of an event is valuable for minimizing exploration cost. When an agent is to explore which events are causes of E = e, if he/she already knows that X = x is cause of E = e and there is a direct causal chain from X=x to E=e, then he/she don't have to actually or virtually test all the possible value settings of relevant variables one by one; the agent can simply start from the position X=x and use the three reasoning operations to explore outward, inward and sideways.

In other words, if the agent has three kind of knowledge: (1) already-known actual causes (e.g., X=x is cause of E=e) ;(2) local causal knowledge (e.g., X=x is Y=y's direct cause or X=x is Z=z's direct effect); (3) direct chain knowledge (e.g., {X,Y,E} or {Z,X,E} is subset of some direct cause chain toward E=e), he/she can calculate whether another event (Y=y or Z=z) is E=e's cause without doing experiments.

---

[10] I guess the connection between interpolation operation and chain feature of causation is a source of people's intuition that a cause in some sense produces a effect: because in causal exploration people can 'conquer' from one event X=x to another event Y=y along the chain, people are inclined to think that X=x produces Y=y.

So we have reason to take the chain feature as a requirement for actual causation:

**Weak Continuity Requirement**: if X = x is the cause of E = e, there is a a chain {X=x, …, E=e} such that each member in the chain is direct cause of its successor.

In most cases, when we say X=x is cause of E=e, the intuition demands not only that X=x belongs to some set $\vec{S} = \vec{s}$ which ensures E=e', but also that X=x belongs to some sets which ensures that E=e is caused in the way or routine that it is actually caused. In other words, it is demanded that there is at least a direct causal chain from X=x to E=e and for each event Z=z in the chain, there is a set $S_z = s_z$ which satisfies: (1) X=x $\in \overrightarrow{S_z} = \overrightarrow{s_z}$ ; (2) $S_z = s_z$ is nonrendundant sufficient set for Z=z. We can formulate a more strict requirement based on this intuition:

**Strong Continuity Requirement**: if X = x is cause of E = e, then there is set {X = x,..., E = e} which satisfies: (1) each member of the set is direct cause of its successor; (2) for any member Z=z in this set, there is a set $\overrightarrow{S_z} = \overrightarrow{s_z}$ , X=x $\in \overrightarrow{S_z} = \overrightarrow{s_z}$, and $\overrightarrow{S_z} = \overrightarrow{s_z}$ is a non-redundant sufficient set for Z=z.

# 5 Risk Cost and Abnormality Requirement of Causation

The phenomena of causal selectivity exists beyond the requirements in previous chapters.[11] We often take some events as key cause of an effect and take other events as background conditions. For instance, When a match is lit in the room, we tend to think movement of the match rather than presence of oxygen is cause of the fire, although in the structural function of the fire, Fire = Match & Oxygen, Match and Oxygen's status is equal. When Jim is eating in the kitchen on the second floor, we generally don't not think the floor's being firm is a cause of Jim's survival, although strictly speaking, if the floor is not firm and cracks, Jim will fall and die.

[11] Halpern, J. Y. and Christopher Hitchcock. [2015]: 'Graded Causation and Defaults', British Journal for Philosophy of Science, 66, p.443.

Causal selectivity can be explained by minimization of risk cost. Suppose $\vec{S} = \vec{s}$ is a nonredundant sufficient set for E=e. If the agent desires E=e but fails to intervene with $\vec{S} = \vec{s}$ or part of $\vec{S} = \vec{s}$ , the agent faces the risk that E=e might not occur, but in the mean time, he/she saves the intervention cost for intervening with $\vec{S} = \vec{s}$ or part of it. If the risk cost is significantly lower than the intervention cost, there is good reason not to intervene with $\vec{S} = \vec{s}$ or its parts.

Quantitively, if we use $\mathrm{Loss}_{\mathrm{risk}}(\vec{S} = \vec{s})$ to denote cost of the total risk that E=e might not occur if intervention with $\vec{S} = \vec{s}$ is absent, use $\mathrm{Loss}_{\mathrm{miss}}\ (\vec{S} = \vec{s})$ to denote the cost of E's nonoccurrence in a particular scenario where $\vec{S} \neq \vec{s}$ and E≠e, use $S_{\mathrm{miss}}$ to denote the set of scenorios where $\vec{S} \neq \vec{s}$ and E≠e, and use P(S) to denote the probability of a particular scenario S, their relationship is as below:

**Formula 5.1, Loss for risk:**

$$\mathrm{Loss}_{\mathrm{risk}}\left(\vec{S} = \vec{s}\,\right) = \mathrm{Loss}_{\mathrm{miss}}\left(\vec{S} = \vec{s}\,\right) * \sum_{s \in S_{\mathrm{miss}}} P(s)$$

Halpern and Hitchcock (2015) proposed a requirement of causal selection for actual causation[12] and the requirement can be seen as a way to filter the events with low risk cost. Briefly speaking, Halpern and Hitchcock's idea is that if $\vec{X} = \vec{x}$ is cause of E=e in scenario ($\vec{u}$, M), then there should be some contrastive scenario S' where: (1) $\vec{X} \neq \vec{x}$ , (2) there is a set of variables $\vec{Z}$ whose members are all on paths from $\vec{X}$ to E, and all members in $\vec{Z} - \vec{X}$ follow their structural function;(3) E≠e and (4) S' is at least as normal as the actual scenario S.

[12] Halpern, J. Y. and Christopher Hitchcock. [2015]: 'Graded Causation and Defaults', British Journal for Philosophy of Science, 66, p.424, 435.

I revised Halpern&Hitchcock's requirement to make it work well with other requirement in previous chapters:

**Definition 5.2** Intrinsic Scenario. If $\vec{X} = \vec{x}$ is an sufficient set for E=e in scenario S=($\vec{u}$,M), then：let $\vec{B} = \vec{b}$ be the set of all $\vec{X} = \vec{x}$ 's member's ancestors. For any memer Y in $\vec{B} - \vec{X}$ , (1) remove Y from the endogenous variable set $\vec{V}$.(2) for any Z in $\vec{V}$ whose structural function's arguments contain Y, replace Y with its value in S in that function's expression. The resulting scenario Si=($\vec{u}$, Mi) is an intrinsic scenario between $\vec{X} = \vec{x}$ and E=e.

**Definition 5.3**. Causal Selection Requirement. If $\vec{S_v} = \vec{s_v}$ is an actual cause set for E=e in ($\vec{u}$, M), then there exist scenario S' where: (1) $\vec{X} \neq \vec{x}$, (2) all variables which are functionally intervened with follow their structural functions; If in the discussed case only reliable scenarios of ($\vec{u}$, M) is relevant, all noninitial variables follow their structural function, (3) E≠e; (4) the intrinsic scenario Si' of S' is at least as normal as the intrinsic scenario Si of S.

The reason for using Intrinsic Scenario is: (1) since it is not specified how a simplified model M's endogenous variables $\vec{V}$ shall be selected, it might be the case that $\vec{V}$ contains many variables which are not relevant to the way how $\vec{X}$ influences E. By creating an intrinsic Scenario, these variables are removed away. (2) the intrinsic scenario procedure helps to avoid some counter intuition application of definition 5.3.

According to the requirement above, if $\vec{X} = \vec{x}$ meets the requirement, there is at least one scenario S ' where intervention with $\vec{X} = \vec{x}$ is absent and the probability of E≠e is no less than the probability of E = e. Thus, at least in scenario S' it is reasonable to intervene with $\vec{X} = \vec{x}$.

In contrast, if $\vec{X} = \vec{x}$ failed to meet the requirement, then for any scenario S' which satisfies ($\vec{u}$,M)|=[ $\vec{X} \neq \vec{x}, \vec{V} - \vec{X} - \vec{F} = (\vec{v} - \vec{x} - \vec{f})'$] E≠e, either S' is less probable than S so that the actual scenario S is no more an abnormal scenario, or the probability of S' is not comparable with that of S so that the agent is unable to judge whether the actual scenario S is an abnormal scenario.

The next question is , how to compare the probability of two scenarios without knowing the specific value of their probability? According to Bayesian network theory, for a causal scenario S containing ordered events $\{X_1 = x_1, \dots, X_n = x_n\}$, the probability of this scenario [13]is:

$$P(X_1 = x_1, \dots, X_n = x_n) = \prod_{xi \in s} \left( P(X_i = x_i | PA_i = pa_i) \right)$$

Where $PA_i = pa_I$ is the value setting of $X_i$'s parent event set in S, i.e., the argument set of $X_i$'s structural function.

When we are to compare the probabilities of two scenarios S = $\{X_1 = x_1, \dots, X_n = x_n\}$, S '= $\{X_1' = x_1', \dots, X_n' = x_n'\}$, both probabilities can be calculated in the way above mentioned:

**Formula 5.4** Probability of Scenario

[13] Pearl, J. [2000]: Causality: Models, Reasoning, and Inference, NY: Cambridge University Press. P.14.

$$P(S) = \prod_{xi \in s} \left( P(X_i = x_i \,|PA_i = pa_i) \right)$$

$$P(S') = \prod_{xi \in s'} \left( P(X_i = x_i' \,|PA_i = pa'_i) \right)$$

In everyday life, we often do not know the exact value of each $P(Xi = xi \,|PA_i = pa_i)$ in the right side of the formulas. However, usually the ordering of $P(Xi = xi \,|PA_i = pa_i)$ and $P(Xi = xi\,'|PA_i = pa'_i)$ is knowable:[14]

(1) Suppose $X_I$ is an initial endogenous variable in the model. If there is some value $x_{i0}$ that for any $x_i$ in $X_i$'s domain, $P(x_{i0}) \geq P(x_i)$, $x_{i0}$ usually would be seen as $X_I$'s default value. Because $PA_i = \{\}$, it can be implied that for any $x_i$ in $X_I$'s domain, $P(x_{i0}|PA_i = \{\}) >= P(x_i|PA_i = \{\})$.

(2) Suppose $X_I$ is an non-initial endogenous variable in the model. Let $f_{xi}(pa_{xi})$ be $X_I$'s value following its structural function. In reliable scenarios, for any $x_i$ in $X_I$'s domain and any value setting of $X_I$'s parents set $pa_{ij}$, $P(f_{xi}(pa_{xij})|pa_i) >= P(xi'|pa_{ij})$.

For two similar scenarios S and S ', if for any endogenous variables Xi ,$P(X_i = x_i \,|PA_i = pa_i) >= P(X_i = x_i' \,|PA_i = pa_i')$, [15] then due to Formula 5.3, P (S)> = P (S ').

Using the procedures above, if we know the default value of each initial variable and the structural function for each non-initial variable, we can compare any two scenarios S and S', and decide whether P (S) $\geq$ P ($S'$), P (S) <P (S '), or they are incomparable.

---

[14] This account has affinity with Halpern and Hitchcock's concept of Default Rule 1, the difference is that in part (1) of my account, exogenous parents are accommodated. See Halpern, J. Y. and Hitchcock, C. [2013]: 'Compact representations of causal models', Cognitive Science, 37, p. 1004.

[15] Here I add apostrophe sign to variable value to mean it is a value setting in S'.

# 6 An Actual Causation Definition and Comparison with Other Definitions

I present a definition of actual causation by incorporating requirements of actual causation in previous chapters.

**Definition 6.1**, Direct cause

X = x is direct cause of Y = y, iff in the scenario S= ($\vec{u}$, (($\overrightarrow{PA_y}$ ∪{Y}, R), F)), X=x satisfies the following conditions:

(1) Sufficiency: X = x ∈ $\vec{S} = \vec{s}$ ; let $\vec{W} = \vec{V} - \vec{S} - \{Y\}$, for any $\vec{W} = \vec{w}'$, ($\vec{u}$, M) $\models [\vec{S} = \vec{s}, \vec{W} = \vec{w}']$ Y = y

(2) Non-redundancy: any subset $\overrightarrow{S'} = \vec{s}$ of $\vec{S} = \vec{s}$ doesn't satisfies (1)

(3) Abnormality: there is some scenario S' where ($\vec{u}$, M) $\models [\vec{S} = \overrightarrow{s'} \neq \vec{s}, \vec{W} = \overrightarrow{w''}]$ $Y \neq y$ , and the probability of S' is not less than that of S.

**Definition 6.2**, Actual Cause

Event C = c is actual cause of E = e in scenario S= ($\vec{u}$, M) iff C=c and E=e satisfies the following conditions:

(1)Sufficiency: there exist three set $\overrightarrow{IS}$, $\overrightarrow{KS}$ , $\vec{W}$, which satisfy:

(a) $\overrightarrow{IS} \subseteq \vec{V}$, $\overrightarrow{KS} \subseteq \vec{V}$, $\vec{W} = \vec{V} - \overrightarrow{IS} - \overrightarrow{KS}$, for some subset W' of W and for any value setting w' of W': ($\vec{u}$, M) $\models [\overrightarrow{IS} = \overrightarrow{is}, \overrightarrow{W'} = \overrightarrow{w'}]$ E = e

(b) If in the case only reliable scenarios of S is relevant, $\overrightarrow{KS}$ is the set of all noninitial endogenous variables. If not, $\overrightarrow{KS}$ ={}.

(2)Non-redundancy: For any $\overrightarrow{IS'} = \overrightarrow{is'} \subseteq \overrightarrow{IS} = \overrightarrow{is}$, $\overrightarrow{FS'} \subseteq \overrightarrow{FS}$, $(\overrightarrow{IS'} = \overrightarrow{is'}, \overrightarrow{FS'})$ doesn't satisfies (1).

(3) Abnormalities: there exist a scenario S' where $(\vec{u}, M) \models [\overrightarrow{IS} = \overrightarrow{is'} \neq \overrightarrow{is}, \overrightarrow{W} = \overrightarrow{w''}]\ E \neq e$ and the probability of the intrinsic scenario Si' of S' is no less than the probability of that of S.

(4) Continuity: there exist a ordered set of {C = c,..., E = e} which satisfies:

(a) each member in the set is direct cause of its successor.

(b) for any member Z=z in the set beside C, C=c and Z=z satisfies (1),(2),(3) .

If intentional action is involved in a causal scenario, the case become a bit complicated. If we use apply Definition 6.2 to intentional action-involving cases, counter-intuitional judgement may appear. For instance:

**Example 6.3**, Weslake's Careful Poisoning. "Assistant Bodyguard puts a harmless antidote in Victim's coffee (A = 1). Buddy then poisons the coffee (B = 1), using a poison that is normally lethal, but which is countered by the antidote. Buddy would not have poisoned the coffee if Assistant had not administered the antidote first. Victim drinks the coffee and survives (D = 0)."[16] Definition 6.2 judges that Buddy's poisoning (B=1) is cause of the victim's not being killed (D=0), yet it is counterintuitional that a counteracted poisoning of the victim is cause of survival of that victim.

**Example 6.4**, Dual Switch Lamp. A lamp is connected with Switch A and Switch B. The lamp is turned on iff one switch is on and the other is off. Jim turns

[16] Weslake, B. (2015). A Partial Theory of Actual Causation. British Journal Philosophy of Science.p22.

Switch A on (J=1). Karl would turns Switch B on only when Switch A is off.(K=0) Thus, the lamp is turned on. (L=1) [17] Definition 6.2 judges that Karl's not turning Switch B on is a cause of the lamp's being on, which is counter intuitional.

We can add a further requirement for intentional action to deal with these problems.

**Requirement 6.5 Requirement for Intentional Action.**

(1) Every intentional action B is composed of some intention event I=I and some following motion event A=a， and I is A's parent variable in the relevant model.

(2) B={I=i, A=a} is cause of event E=e, iff both I=i and A=a are causes of E=e.

With this requirement, in Example 6.3 Assistant and Buddy's intentions should be added to the model as variables. Thus ,the scenario's structural function set becomes {AI=1;BI=1;A=AI;B=BI&A; E=~A&B}. Neither Assistant's intentional action {AI=1;A=1} or Buddy's intentional action {BI=1; B=1} meet the Intentional Action Requirement. This result is consistent with the intuition that neither Assistant nor Buddy's action is cause of the victim's survival. Similarly, in Example 6.4 Jim and Karl's intentions should be added to the model as variables, and the scenario's structural function set becomes {JI=1;KI=1; J=JI;K=KI&~J; E=(A&~B)|(~A&B)}. Only Jim's intentional action {JI=1, J=1} meet the Intentional Action Requirement, and this result is consistent with the intuition that Jim's action rather than Karl's causes the lamp's being on. Other causal cases with intentional actions can also be dealt in this way.

[17] See Giordani,A. (2016). An Internal Limit of the Structural Analysis of Causation, Axiomathes, 26: p444, "world1".

I collect 66 different causal cases from actual causation literature. I compare the intuition of causation in these cases with applications of three recent definition of actual causation using causal modelling approach: (1) Definition 6.2 with the intentional action requirement; (2) Halpern and Hitchcock's definition for graded causes [18]; (3) Weslake's partial theory definition for actual causation[19]. The complete result of comparison is in Appendix II. A summary of the comparison is as below:

**Table 6.6 Comparison of Definition Applications**

| | counterintuition cases count | total cases count |
|---|---|---|
| Halpern&Hitchcock | 9 or 18[20] | 66 |
| Weslake | 27 | 66 |
| Liang | 0 | 66 |

Specifically, Comparing definition 6.2 with Halpern and Hitchcock's definition, there are three types of cases that worth mentioning.

[18] Halpern, J. Y. and Christopher Hitchcock. [2015]: 'Graded Causation and Defaults', British Journal for Philosophy of Science, 66, p.424, 435.

[19] Weslake, B. [2015]: 'A partial theory of actual causation', British Journal for Philosophy of Science, Forthcoming. p.23.

[20] If we consider cause by omission as cause intuitively, then the count of counterintuition cases for Halpern&Hitchcock definition is 18. If we don't consider cause by omission as genuine cause, the count is 9.

(1) Non-binary domains. When the event variable's domain size is more than 2, sometimes HPH definition would yield counterintuitive judgement. Definition 6.2 doesn't.

(2) Variables carrying multiple properties. When one event variable has 2 or more causally relevant property(such as intensity and color), sometimes HPH definition yields counterintuitive judgements. Definition 6.2 doesn't.

(3) Cause by omissions. It is claimed that the Halpern and Hitchcock definition can explain why some think omission is not genuine cause while some others think omission is[21]. Due to Halpern&Hitchcock's theory, this is because "those who maintain that omissions are never causes can be understood as having a normality ranking where absences or omissions are more typical than positive events"[22], while some others don't endorse such a normality ranking.

In applications of Definition 6.2, many omissions refuted by Halpern&Hitchcock definition are seen as cause. Then a question is , does my theory have the potential to explain why people have intuitive disagreement concerning cause by omission?

From the view point of cost minimization, the disagreement above reflect such a question: when we are to consider whether to cancel intervening with some events to minimize intervention cost, should the object of the canceled intervention be an entire sufficient set or some single event in a sufficient set ?

In definition 6.2, the object of the canceled intervention is a sufficient set, i.e., we only consider whether cancelling intervention of such a set is desirable for cost

[21] Halpern, J. Y. and Christopher Hitchcock. [2015]: 'Graded Causation and Defaults', British Journal for Philosophy of Science, 66, p.437.

[22] Ibid., p.437.

minimization. If we make some revision to definition 6.2 and change the object of canceled intervention to some single event, the Abnormality Condition of definition 6.2 becomes:

(3') Abnormalities: there exist a scenario S' where $(\vec{u}, M) \models [C = c' \neq c, \vec{W} = \overrightarrow{w''}]\ E \neq e$ and the probability of the intrinsic scenario Si' of S' is no less than the probability of that of S.

In the alternative of Definition 6.2 by replacing (3) with (3'), generally omissions will no longer be considered cause. Therefore, my actual causation theory can also explain people's intuitive disagreement in cause by omission. But according to my theory, the essence of the disagreement is disagreement in how to think of the object of canceled intervention, i.e., disagreement in whether the proper object should be a sufficient set of the effect event or a single event in such a set.

# Appendix I: Proof of Interpolation Operation Principle

**Proposition:** Suppose $\vec{S} = \vec{s}$ is a cause net of E = e in a reliable scenario $(\vec{u}, M)$ .

Replace some member X=x of $\vec{S} = \vec{s}$ with a set $\vec{H} = \vec{h}$, and $\vec{H} = \vec{h}$ should satisfy: for each direct cause chain {X=x, …, E=e} where each chain member is a direct cause of its successor, the successor of X=x is included in $\vec{H} = \vec{h}$. The new set after replacement $\vec{S'} = \vec{s'}$ is an sufficient set in reliable scenarios.

**Proof**:

1, Partition $\overrightarrow{S_{dir}} = \overrightarrow{s_{dir}}$ into $\vec{S_1} = \vec{s_1} \subseteq \vec{S} = \vec{s}$ and $\vec{S_2} = \vec{s_2} \subseteq \overrightarrow{S_{ex}} - \vec{S} = \overrightarrow{s_{ex}} - \vec{s}$ .Let the expression 'A ensure B' be short for A is a sufficient set for B.

2, if X=x is direct cause of E=e, then E = e will be added into $\vec{S_1} = \vec{s_1}$, and {E = e} ensure its own occurrence.

3, if X=x is not direct cause of E=e, then:

3.1, , $\vec{S'} = \vec{s'}$ ensures $\vec{S_1} = \vec{s_1}$ . Because, since X = x is not member of $\overrightarrow{S_{dir}} = \overrightarrow{s_{dir}}$ , the relation $\vec{S_1} = \vec{s_1} \subseteq \vec{S'} = \vec{s'}$ would still hold after deleting X=x from $\vec{S'} = \vec{s'}$.

3.2, $\vec{S'} = \vec{s'}$ ensures $\overrightarrow{S_{ex}} - \vec{S} = \overrightarrow{s_{ex}} - \vec{s}$, hence $\vec{S'} = \vec{s'}$ ensures $\vec{S_2} = \vec{s_2}$. This is because,for any variable Z=z in $\overrightarrow{S_{ex}} - \vec{S}$, either X=x is an ancestor in some direc cause chain to Z=z or not, $\vec{S'} = \vec{s'}$ ensures Z=z :

3.2.1, if X=x is an ancestor in some direct cause chains to Z=z, then for any of these chains, during the interpolate operation either Z or some variable between X=x and Z=z in the chain is added into $\vec{S'} = \vec{s'}$, and it can be recursively proved that $\vec{S'} = \vec{s'}$ ensures Z=z.

3.2.2, if X=x is not an ancestor in some direct cause chain to Z=z, then the value of Z won't be influenced by X=x's being removed.

4 Combining 2 and 3, $\overrightarrow{S'} = \overrightarrow{s'}$ ensures E = e.

# Appendix II Comparison OF Definition Application By Cases

In the Table below I collect 66 causal scenarios from different souces and compare three definitions' application to these scenarios.

The Formulas column shows the structural functions of each scenario, where e means the effect event. If a non-effect variable Y's domain size is larger than 2, I explicitly show its domans in the formula in forms like "${'y':{0,1,2}}". If an event variable X carries two properties A and B(such as intensity and color), I use one integar x= (Property_B_Domain_Size * Property_A_Value + Property_B_Value) to denote the value of X.

The Intuition column shows what our main intuition judges as causes of the effect e. The Liang, HPH, Weslake column shows what each author's definition judges as causes of e.[1] If the content in a Liang, HPH or Weslake unit matchs intuition, the unit is left blank.

The Source column shows the source of citation of each scenaro:

P stands for scenarios in Paul,L.A.and N.Hall(2013).Causation:A User's Guide.Oxford,U.K.:Oxford University Press. The following number means original figure number.

G stands for scenarios in Giordani,A. (2016). An Internal Limit of the Structural Analysis of Causation, Axiomathes, 26: pp429-450. The following number means page number.

W stands for scenarios in Weslak,B. (2015). A Partial Theory of Actual Causation. British Journal for the Philosophy of Science. To Appear. The following number means original example number.

| | Liang | HPH | Weslake | Intuition | Formulas | Source |
|---|---|---|---|---|---|---|
| 1 | | | d | c,d | a=1;c=1;d=c;b=~c&a;e=d\|b | P1 |
| 2 | | | | a,b | a=1;c=0;d=c;b=~c&a;e=d\|b | P2 |
| 3 | | c*[2] | | a,c | a=0;c=1;e = ~a & c | P3 |
| 4 | | | | c,d | c=1;d = c;b = c;e=d | P8 |
| 5 | | | | f,c | f=1;d=1;c=f;e= c;a=d;b=a | P9 |

[1] Liang Definition is Definition 6.2 of this paper. HPH Definition is the definition from Halpern, J. Y. and Christopher Hitchcock. [2015]: 'Graded Causation and Defaults', British Journal for Philosophy of Science, 66, p.424, p.435. Weslake Definition is from Weslak,B. (2015). A Partial Theory of Actual Causation. British Journal for the Philosophy of Science,p23.

[2] I use asterisk to mark the counterintuitive scenarios which would be intuitive if all causes by omissions are not considered as genuine cause.

| | | | | | | |
|---|---|---|---|---|---|---|
| 6 | | a* | | a,d,f | a=1;c=1;b = c;d= c;f = ~d & b;e = ~f &a | P10 |
| 7 | | | | c,a | c=1;a=1;e=c\|a | P11 |
| 8 | | | d | c,d | a=1;c=1;f=1;b= ~c &a;g = ~c &f;d = c;e = b \| g \| d | P12a |
| 9 | | | | a,b,f,g | a=1;c=0;f=1;b= ~c &a;g = ~c &f;d = c;e = b \| g \| d | P12b |
| 10 | | | | D | a=1;c=1;f=1;b= ~c &a;g = ~c &f;d = 1;e = b \| g \| d | P13a |
| 11 | | a,d,b,g,f* | a,b,f | a,c,f,b,d,g | a=1;c=0;f=1;b= ~c &a;g = ~c &f;d = 1;e = b \| g \| d | P13b |
| 12 | | a,b | | a,b,c,f | a=1;c=1;b=a;f=c;d=a&~c;e=(b&d)\|(d&f)\|(b&f) | P14a |
| 13 | | | | c | c=1;e=c | P20 |
| 14 | | c* | | c,f | c=1;f=0;d=0; b=c&f; e = d\|(c&~f) | P21a |
| 15 | | | | d | c=1;f=1;d=1; b=c&f; e = d\|(c&~f) | P21b |
| 16 | | | | c | c=1;d=1;e=~c&d | P22 |
| 17 | | | | a | a=1;b=1;c=1; e= a\|(b&~c) | P23 |
| 18 | | | | {} | c=0;g=0;i=0;d=0; e=(~c&d)\|(c&g)\|(c&~i&~g&d) | P25 |
| 19 | | | d,g | c, d, g | c=1;g=1;i=1;d=1; e=(~c&d)\|(c&g)\|(c&~i&~g&d) | P26 |
| 20 | | i,c* | | i, c, g | c=1;g=0;i=1;d=1; e=(~c&d)\|(c&g)\|(c&~i&~g&d) | P27 |
| 21 | | | | a,c,d | a=1;b=1;c=1;d=b&~c;e=a&~d | P29 |
| 22 | | | d | c,g,d or c,g,d,f | c=1;a=1;f=0;g=c&~f;d=g;b=a&~g;e=d\|b | P38 |
| 23 | | c* | | c,d | c=1;a=0;b=0;d=a\|b;e=~d&c | P39 |
| 24 | | a* | | a, d | a=1;c=1;d=0 ; e= a&~d | P40 |
| 25 | | a* | | a,d,f | a=1;g=0;c=1;b=g\|c;d=c;f=b&~d;e=a&~f | P41 |
| 26 | | | a,d,f | a,c,d,f | a=1;g=1;c=1;b=g&~c;d=c;f=b\|d;e=a\|f | P42 |
| 27 | | | d,g | c,d,g | f=1;c=1;a=1;d=c;b=~c&a;g=d\|b;e=~g&f | P43 |
| 28 | | | | h,c,d,i,g | h=1;f=1;c=1;a=1;d=c;b=~c&a;i=d\|b;g=~i&f;e=~g&h | P44 |
| 29 | | | a,d | a,b,d | a=1;c=1;b= { 1 if a&c,2 if a&~c,0 if~a};d=(b==1);f=(b==2);e=d\|f${'b':{0,1,2}} | P45 |
| 30 | | | a,f | a,b,f | a=1;c=0;b= { 1 if a&c,2 if a&~c,0 if~a};d=(b==1);f=(b==2);e=d\|f${'b':{0,1,2}} | P46 |
| 31 | | | | a,b,c,d | a=1;c=1;b=a;d=c&b;f=~c&b;e=d | P47 |
| 32 | | | | a,b,d | a=1;c=0;b=a;f=c;d=a&~c;e=(b&d)\|(d&f)\|(b&f) | P14b |
| 33 | | | | a,b | c=1;a=1;f=1;g=c&~f;d=g;b=a&~g;e=d\|b | P38b |
| 34 | | | a,b | [] | a=0;b=1;e= a&~b | G12.world2 |
| 35 | | | | [] | a=0;b=a;e=a&~b | G13.world2 |
| 36 | | | b | a,f | f=1;a=f; g=1; b=g&~a; e = (a&~b)\|(~a&b) | G16.world1 |
| 37 | | | | b,g | f=0;a=f; g=1; b=g&~a; e = (a&~b)\|(~a&b) | G16.world2 |
| 38 | | | | b,g | f=1;a=f; g=1; b=g&a; e = (a&~b)\|(~a&b) | G17.world1 |
| 39 | | | b | [] | f=0;a=f; g=1; b=g&a ;e = (a&~b)\|(~a&b) | G17.world2 |
| 40 | | a* | a,c,b,d | a,d | a=1;b=0;c=1;d= b&~c;e=a&~d | G19.e4.4b |
| 41 | | a | a | a,c | a=1;c=1;d=a&~c;e=(a&d)\|(d&c)\|(a&c) | P14c |
| 42 | | | b,f | b,c,f | a=1;b=1;c=1;d=a&~c;f=b&c;e=d\|f | |
| 43 | | {} | | b | c=1;a=~c;b=c;e= a\|b | W.e6 |
| 44 | | | | b,g | f=1;a=f; g=1; b=g&a; e=(a&b)\|(~a&~b) | W.e7 |
| 45 | | | {} | c,f | f=1;c=f;g=1;a=g&~c;e=a\|c | W.e1 |
| 46 | | | {} | c,d | a=1;c=1;d=c;b= a&~d; e=b\|d | W.e3 |
| 47 | | | | c | a=1;b=0;c=1;e=(a&b)\|c | W.e5 |
| 48 | | | | {} | f=1;g=1;a=f;b= g&a;e=~a&b | W.e11 |
| 49 | | c,m | {} | m | m=1;c=1;e= (m==1)\|(c&(m!=2))${'m':{0,1,2}} | W.e8 |
| 50 | | | {} | v,w,m | v=1; w=1 ; m=v+w ; e=(m>=1) ${'m':{0,1,2}} | W.e4 |
| 51 | | a,c,b | {} | b,c | a=1;b=-1;c=-1;e=(a==b)\|(b==c)\|(a==c) $ {'a':{-1,0,1},'b':{-1,0,1},'c':{-1,0,1}} | W.e9 |
| 52 | | a,c,b,f | a | a,c,f | a=1;b=1;c=1;f={2 if a&c, 1 if b&~c, 0 if~((a&c)\|(b&~c))}; e = (f>0)$ {'f':{0,1,2}} | W.e12 |
| 53 | | a,b | a,d | a,b,c,d | c=5;a=10;b=a-c;d = c;e =(d+b>=10)${'a':{0,10}, 'c':{0,5},'d':{0,5}, 'b':{-5, 0, 5, 10}} | P15 |
| 54 | | | a,c,b,d | a,b,c,d | c=10;a=10;d=c-a/2;b=a-c/2;e=(b+d>=10)${ 'a':{0,10},'c':{0,10},'b':{-5,0,5,10},'d':{-5,0,5,10}} | P16 |
| 55 | | | | a,b | c=10;a=10;b=a-c/2;e=(b>0)$ {'a':{0,10},'c':{0,10},'b':{-5,0,5}} | P17 |
| 56 | | | | c | c=7;d=4;e = {0 if c==0&d==0, 3+c%3 if c/3>=d/3, 3+d%3 if c/3<d/3} ${'c': {0,7,8}, 'd': {0,4,5}} | P34 |
| 57 | | c,d | {} | c | c=7;d=4;e = {0 if c==0&d==0, 3+c%3 if c/3>=d/3, 3+d%3 if c/3<d/3} ${'c': {0,4,5,7,8}, 'd': {0,4,5,7,8}} | P34b |
| 58 | | | | c | c=8;d=5;e = {0 if c==0&d==0, 3+c%3 if c/3>=d/3, 3+d%3 if c/3<d/3} ${'c': {0,7,8}, 'd': {0,4,5}} | P37 |
| 59 | | | | c | c=9;d=5;e={c if c>0, d+4 if c==0&d>0, 0 if c==0&d==0}; $ {'c':{0,9,10,11},'d':{0,5,6,7}} | P30 |
| 60 | | | {} | c,f | c=9;d=5;f= {((c-c%4)/4-(d-d%4)/4)*4+c%4 if c/4>d/4, 0 if c/4<=d/4};<br>g={((d-d%4)/4-(c-c%4)/4)*4+d%4 if c/4<=d/4, 0 if c/4>d/4};<br>e={8+f%4 if f/4>=g/4, 8+g%4@f/4<g/4};<br>$ {'c':{0,9,10,11},'d':{0,5,6,7},'f':{0,5,6,7,9,10,11},'g':{0,5,6,7}} | **P31** |
| 61 | | a,c,d | a | a,c | c=7;d=4;a= {c if c/3>=d/3, d if c/3<d/3};b={d if c/3>=d/3, c if c/3<d/3};f=3+b%3;<br>e= 3+a%3 ${'c':{0,6,7,8},'d':{0,3,4,5},'a':{0,6,7,8},'b':{0,3,4,5}} | **P32** |
| 62 | | | | d | c=9;d=5;h={0 if c*d==0, 1 if (c*d!=0)&((c%4!=d%4)\|(c/4<d/4)), 2 if (c*d!=0)&( c/4>=d/4) &(c%4==d%4)};<br>b = d* (1 -h%2);a= c*(1-h/2);g = {(b/4-a/4)*4+b%4 if b/4>a/4, 0 if b/4<=a/4};<br>f ={(a/4-b/4)*4+a%4 if a/4>b/4,0 if a/4<=b/4};e = {8+f%4 if f/4>=g/4, 8+g%4 if f/4<g/4}<br>${'c':{0,9,10,11},'d':{0,5,6,7},'h':{0,1,2},'a':{0,9,10,11},'b':{0,5,6,7},'f':{0,5,6,7,9,10,11},'g':{0,5,6,7}} | **P33** |
| 63 | | | | c | c =7; e=6+c%3 ${'c': {0,7,8}} | P35 |
| 64 | | | | d | d =4; e=6+d%3 ${'d': {0,4,5}} | P36 |
| 65 | | | | a,c,b | a=15;b=1;c= (a/15)*(15+b);d=c/15;e=(c%15)==1 $ {'a':{0,15},'c':{0,15,16}} | P48 |
| 66 | | | {} | c,b | a=15;b=2;c= (a/15)*(15+b);d=c/15;e=(c%15)==1 $ {'a':{0,15},'c':{0,15,16}} | P49 |